%% file: root.tex
\documentclass[letterpaper, 10 pt, conference]{ieeeconf}  % Comment this line out if you need a4paper

\IEEEoverridecommandlockouts                              % This command is only needed if
\usepackage{graphics} % for pdf, bitmapped graphics files
\usepackage{epsfig} % for postscript graphics files
\usepackage{amsmath} % assumes amsmath package installed
\usepackage{amssymb}  % assumes amsmath package installed
\usepackage{multicol}
\usepackage[bookmarks=true]{hyperref}
\usepackage{xcolor}
\usepackage{booktabs}
\usepackage{graphicx}
\usepackage{svg}
\usepackage{subcaption}

\usepackage[backend=bibtex,style=ieee,natbib=true,mincitenames=1,maxcitenames=2]{biblatex}
\newcommand{\ie}{\textit{i}.\textit{e}., }
\newcommand{\eg}{\textit{e}.\textit{g}. }

\renewcommand{\arraystretch}{1.1}

\title{\LARGE \bf
Proprioceptive State Estimation of Legged Robots with Kinematic Chain Modeling
}

\author{Varun Agrawal$^{1}$, Sylvain Bertrand$^{2}$, Robert Griffin$^{2,3}$, Frank Dellaert$^{1}$% <-this % stops a space
    \thanks{$^{1}$Institute for Robotics and Intelligent Machines and School of Interactive Computing, Georgia Institute of Technology, Atlanta, GA 30332 USA.
        {\tt\small \{varunagrawal,frank.dellaert\}@gatech.edu}}%
    \thanks{$^{2}$Florida Institute for Human \& Machine Cognition, Pensacola, FL 32502 USA.
    	{\tt\small \{sbertrand,rgriffin\}@ihmc.org}}%
    \thanks{$^{3}$Intelligent Systems and Robots, University of West Florida, Pensacola, FL 32503 USA.}
}

\begin{document}

\maketitle
\thispagestyle{empty}
\pagestyle{empty}

%%%%%%%%%%%%%%%%%%%%%%%%%%%%%%%%%%%%%%%%%%%%%%%%%%%%%%%%%%%%%%%%%%%%%%%%%%%%%%%%
\begin{abstract}
Legged robot locomotion is a challenging task due to a myriad of sub-problems, such as the hybrid dynamics of foot contact and the effects of the desired gait on the terrain. Accurate and efficient state estimation of the floating base and the feet joints can help alleviate much of these issues by providing feedback information to robot controllers. Current state estimation methods are highly reliant on a conjunction of visual and inertial measurements to provide real-time estimates, thus being handicapped in perceptually poor environments. In this work, we show that by leveraging the kinematic chain model of the robot via a factor graph formulation, we can perform state estimation of the base and the leg joints using primarily proprioceptive inertial data. We perform state estimation using a combination of preintegrated IMU measurements, forward kinematic computations, and contact detections in a factor-graph based framework, allowing our state estimate to be constrained by the robot model. Experimental results in simulation and on hardware show that our approach out-performs current proprioceptive state estimation methods by 27\% on average, while being generalizable to a variety of legged robot platforms. We demonstrate our results both quantitatively and qualitatively on a wide variety of trajectories.
\end{abstract}

\input{01_introduction}
\input{02_related_work}
\input{03_preliminaries}
\input{04_approach}
\input{05_results}
\input{06_conclusion}

\newpage
\printbibliography

\end{document}

%% file: 01_introduction.tex
%%%%%%%%%%%%%%%%%%%%%%%%%%%%%%%%%%%%%%%%%%%%%%%%%%%%%%%%%%%%%%%%%%%%%%%%%%%%%%%%
\section{INTRODUCTION}

%\paragraph{Problem is important}
%IMU integration for state estimation is an important problem in the area of legged robotics.

Legged robots are considered advantageous in comparison to traditional wheeled robots due to their ability to traverse uneven~\cite{Grandia19iros,Kumar21rss} and unstructured terrains~\cite{Kolvenback20jfr}. 
Deployment of robots in the real-world is difficult since the world is designed to be human-centric. Environments such as multi-level buildings, and unstable or discontinuous ground~\cite{Lebastard11tro_biped_orientation,Semini11bioinsp_hyq} are better tackled by legged robots.
Humanoids, in particular, are well suited for tasks which require human levels of dexterity, manipulation, and characterization~\cite{Breazeal03ijhcs_emotion_humanoids,Hirai98icra_honda_humanoid,Nelson12jrsj_petman}. 

Many challenges with legged robot locomotion and stability can be addressed with accurate and efficient state estimation. The robot's high degree of freedom from its articulated legs, the hybrid dynamics due to discrete foot-contacts, the desired gait specifications, and managing foot slip~\cite{Johnson15jfr_ihmc} are some of the key challenges which can be tackled by accurately estimating the state of the robot's floating base and the trajectory of foot poses. In particular, high frequency state feedback controllers require even higher rate state estimates, which is necessary to deal with uncertain events such as foot slip and balance.

\begin{figure}[t]
	\captionsetup{font=footnotesize}
	\centering

	\begin{subfigure}{0.2\textwidth}
		\label{fig:atlasrobot}
		\includegraphics[width=\textwidth,bb=0 0 601 638]{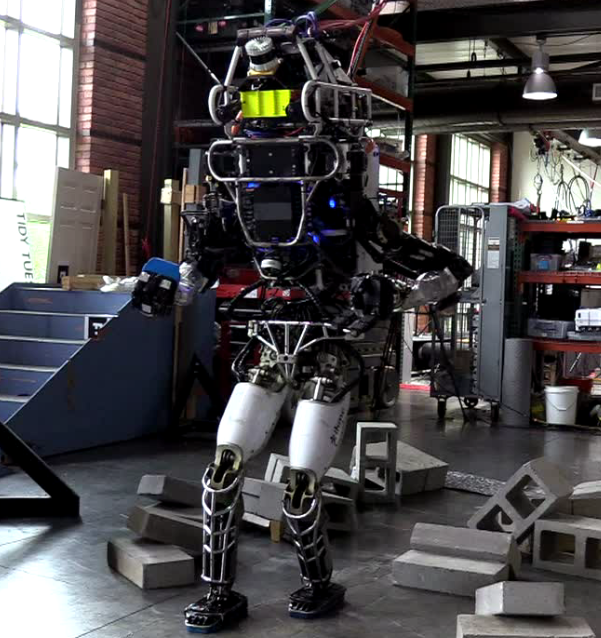}
		\caption{The DRC-era Atlas humanoid by Boston Dynamics in action.}
	\end{subfigure}
	\begin{subfigure}{0.25\textwidth}
		\label{fig:atlas_traj_example}
		\includegraphics[width=\textwidth]{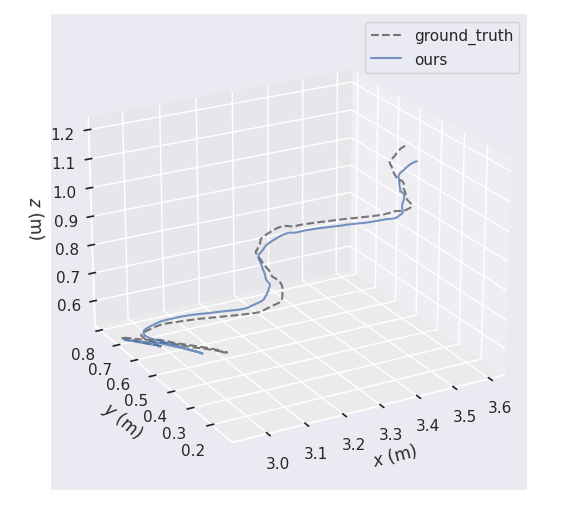}
		\caption{The estimated trajectory of the Atlas walking through obstacles.}
	\end{subfigure}
	\begin{subfigure}{0.23\textwidth}
		\label{fig:atlas_xyz_example}
		\includegraphics[width=\textwidth]{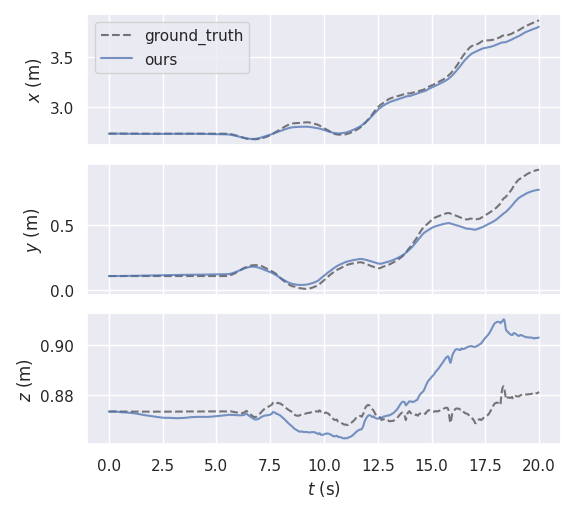}
		\caption{The estimated translation values versus the ground truth.}
	\end{subfigure}
		\begin{subfigure}{0.23\textwidth}
		\label{fig:atlas_rpz_example}
		\includegraphics[width=\textwidth]{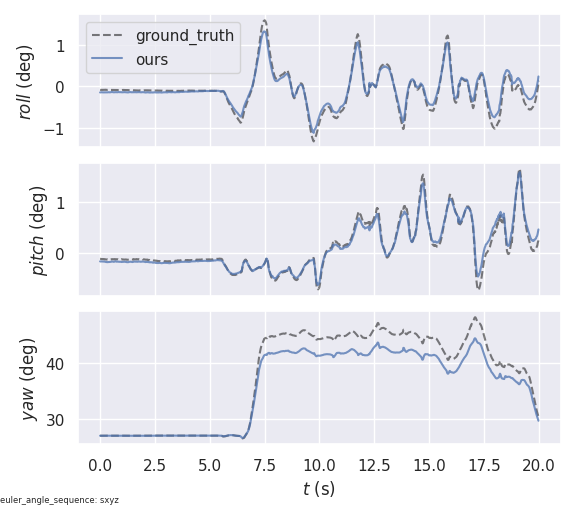}
		\caption{The estimated rotation values versus the ground truth.}
	\end{subfigure}
	
	\caption{A representative example of our state estimator working for the Atlas robot. Experiments are conducted on the A1 robot in simulation and on real hardware for the Anymal C robot by ANYbotics AG and the Atlas, demonstrating the generality of our approach.}

    \vspace{-2em}
\end{figure}

%\paragraph{Not solved}
Current legged robot state estimators leverage exteroceptive sensors which fail in poor environmental conditions and limit the frequency of controllers.
Exteroceptive sensors, such as cameras and LIDARs, are used in conjunction with proprioceptive sensors such as Inertial Measurement Units (IMUs) and joint encoders~\cite{Wisth21tro_vilens}~\cite{Camurri20frontiers_pronto}~\cite{Hartley18iros_vic}, and these methods are reliant on being able to detect and process features in the surroundings.
This requires well-lit, opaque, and structured environments, which is not always the case, while also being limited by the frame rate of visual sensors (\eg cameras), which tend to be an order of magnitude slower than that of IMUs, limiting the publishing rate of state estimates, and subsequently the controller.

%\paragraph{Our contribution}
Proprioceptive state estimation has shown to be a viable solution~\cite{Bloesch12rss_se_imu_leg_kinematics,Bledt18iros,Kuindersma16ar_atlas}, however it suffers from a fair number of problems.
While having a multitude of advantages, such as IMU-rate (\eg 200 Hz) estimates for dynamic control, operating in perception-denied environments~\cite{Tranzatto22arxiv_cerebrus}, and freeing up sensors (e.g. cameras) for other tasks such as mapping~\cite{Camurri20frontiers_pronto}, they suffer from long-term drift (particularly for filtering based approaches~\cite{Bloesch12rss_se_imu_leg_kinematics}), vary for different legged robot platforms, and assume one noise model from the base to the foot without taking into account complex joint interactions and link deflections introduced over time.

We propose a framework to perform kinematic state estimation for legged robots using factor graphs which tackles both issues of drift and generalizability. Using proprioceptive measurements with the kinematic chain of the robot's legs in contact, available from the Unified Robot Description Format (URDF) file, we can directly model the full kinematics of the robot at a time instance with a factor graph, and perform probabilistic inference for the full state of the robot.  Moreover, the factor graphs allow for a generic sensor-fusion mechanism, allowing the incorporation of additional factors and sensors such as leg IMUs, contact sensors, and slip-detectors, depending on the situation, as well as an extensible framework for asynchronous data processing.

Our method consistently produces better state estimates than the baseline, in terms of Absolute Pose Error (APE) and Relative Pose Error (RPE), on a variety of trajectories for each platform.
We demonstrate our results in simulation using the Unitree A1 robot~\cite{Peng20rss, Kumar21rss}, and on physical hardware using the Anymal C quadruped~\cite{Wisth20icra_contact_fg} and the Boston Dynamics' DRC-era Atlas humanoid~\cite{Johnson15jfr_ihmc}. As a comparative baseline, we use a factor graph based implementation of~\citet{Bloesch13iros_se_slippery_uneven}. We choose this baseline since it illustrates the benefit of our approach directly in a fair comparison, while being orthogonal to more recent advances.

%% file: 02_related_work.tex
%%%%%%%%%%%%%%%%%%%%%%%%%%%%%%%%%%%%%%%%%%%%%%%%%%%%%%%%%%%%%%%%%%%%%%%%%%%%%%%%
\section{RELATED WORK}

Use of inertial measurement units (IMUs) for proprioceptive state estimation has a rich history~\cite{Braun13aiaa}. The initial application was for the control of intercontinental ballistic missiles~\cite{Bezick10report_icbm}. Being mission-critical, they are highly sophisticated (tactical or navigation grade) but also very expensive~\cite[Table 1]{Bezick10report_icbm}, making them typically unsuitable for general robotics applications.

More recently, IMUs have seen widespread adoption in the areas of visual-inertial odometry and robot localization~\cite{Leutenegger15ijrr,MurArtal16,Qin18tro_vins_mono}. Initial work focused on sensor fusion based on Kalman filtering~\cite{Bloesch15iros,Mourikis07icra,Usenko16icra,Weiss12icra}.
Recent smoothing methods, such as \citet{Forster17tro}, leverage preintegrated IMU measurements with camera-based visual measurements to perform robot localization~\cite{Eckenhoff19ijrr} and state estimation~\cite{Nisar19rss_vimo}. While these are synergistic with goals such as mapping, they suffer from issues such as lighting dependency, potential lack of visual features, and increased computational cost, causing up to two orders of magnitude slower processing rates.

For legged robots, initial work with proprioceptive state estimation was performed using Kalman filtering and the contact points of the robot with its environment~\cite{Bloesch16humanoids_sense_of_balance,Bloesch13iros_se_slippery_uneven,Bloesch12rss_se_imu_leg_kinematics}. These foundational works focus on using the forward kinematics of the legs to define constraints on the position and velocity of the contact points and thus constrain the biases in the IMU measurements. The major drawback is that the yaw angle (around gravity) and the absolute position of the robot become unobservable, leading to drift errors over long durations. Related to our work,~\citet{Rotella14iros} perform filtering based state estimation for humanoids specifically, using ideas from~\cite{Bloesch13iros_se_slippery_uneven} while leveraging the flat-foot constraint of these robots to tackle yaw angle unobservability.

Recently, there has been an increased interest in smoothing based methods for legged robot state estimation. \citet{Wisth19ral_legged_fg} demonstrate excellent results for state estimation by fusing proprioceptive and exteroceptive sensor measurements in a factor-graph based framework for a quadruped. However, for leg odometry, they rely on interpolation over the states from a Two-State Implicit Filter~\cite{Bloesch17ral_two_state} rather than jointly estimate it.
Closely related to our work,~\citet{Hartley18icra_contact_factors} leverage a factor-graph smoothing framework to jointly optimize over IMU measurements, leg kinematics and contact measurements preintegrated over each stance phase of biped locomotion. A follow up work~\cite{Hartley18iros_vic} shows the promising potential for hybrid system modeling of contacts inherent to bipedal walking. By adding these hybrid preintegrated contact factors with visual features from a stereo camera setup, they are able to demonstrate state estimation over a variety of conditions and terrains. 

%% file: 03_preliminaries.tex
\section{PRELIMINARIES}

\newcommand{\R}{\mathrm{R}}
\newcommand{\position}{\mathrm{p}}
\newcommand{\vel}{\mathrm{v}}
\newcommand{\bgyr}{\mathrm{b}^{g}}
\newcommand{\bacc}{\mathrm{b}^{a}}

In this section, we present some preliminaries on factor graphs, the robot kinematic model, and IMU preintegration in the context of smoothing, all of which serve as the basis for the discussion of our work in subsequent sections.

\subsection{Factor Graph Based Smoothing}
% Smoothing problem
A factor graph is a probabilistic graph model which can be used to express the estimation problem in a simple and elegant manner. It is a bipartite graph represented as $\textit{G} = \{ \phi, \Theta, \varepsilon \}$, where $\phi$ are the factor nodes, $\Theta$ are the variable nodes to be estimated, and $\varepsilon$ are the edges between them.

The factor graph $\textit{G}$ represents the probability density over the variables to be estimated $\Theta$ given the measurement values $Z$, and provides the factorization as
\begin{equation}
P(\Theta | Z) = \prod_{i=1}^{|\textit{G}|}\phi_i(\Theta_i),
\label{factor-graph}
\end{equation}
with each factor $\phi_i(\Theta_i) = P(\Theta_i | Z_i)$ specifying the relationship between its connected subset of variables $\Theta_i \subseteq \Theta$.

Given the factor graph $\textit{G}$, the objective is to compute the \textit{maximum a posteriori} estimate of variables $\Theta$, given as

\begin{equation}
\Theta^* = \arg \max_\Theta \prod_{i=1}^{|\textit{G}|}\phi_i(\Theta_i) = \arg \min_\Theta -\sum_{i=1}^{|\textit{G}|}\text{log} \phi_i(\Theta_i).
\end{equation}

%\subsection{On-Manifold Optimization}

%\subsection{Kinematics Model}
%
%To acquire the kinematics model of any particular robot, we defer to the Universal Robot Description Format (URDF) file.
%The URDF file contains the full specification of the robot's links and joints, which allows the robot to be used in modeling and simulation with high fidelity.
%This specification provides us with the necessary information required to perform full kinematic modeling of the legs (considered as open-chains~\cite{Lynch17book_robotics}), allowing efficient computation of the forward kinematics given the instantaneous joint angle measurements.

\subsection{IMU Preintegration}
In this section, we provide a brief overview of IMU preintegration and how it is used in the factor graph framework. For more details, the reader is referred to the exposition in~\citet{Forster17tro}.

It is important to handle an IMU's high measurement frequency so as to not overwhelm the state estimator when performing smoothing. This is done by forward integrating the IMU's measurements (angular velocity $\omega$ and linear acceleration $a$) across successive $\Delta t$ between two specific timestamps to obtain a summarized compound measurement for the body $B$ in an arbitrary world frame $W$.
%\begin{align}
%	\begin{split}
%	\R^{W}_B(t + \Delta t) = \R^W_B(t) \text{Exp}\left(\int_{t}^{t+\Delta t}\omega^W_B(\tau)d\tau \right)
%	\\
%	\vel^{W}(t + \Delta t) = \vel^W(t) + \int_{t}^{t+\Delta t}a^W(\tau)d\tau
%	\\
%	\position^{W}(t + \Delta t) = \position^W(t) +  \vel^W(t)\Delta t + \int \int_{t}^{t+\Delta t}a^W(\tau)d\tau^2
%	\end{split}
%\end{align}
To prevent the forward integration being performed repeatedly at each new linearization point, a \textit{relative} motion model is computed between the states at time $t_i$ and time $t_j$
\begin{align}
	\begin{split}
		& \Delta \R_{ij} \triangleq \prod_{k=i}^{j-1}\text{Exp} \left( \left( \omega_k - \bgyr_k - \eta^{gd}_k \right)\Delta t \right),
		\\
		& \Delta \vel_{ij} \triangleq \sum_{k=i}^{j-1}\Delta \R_{ik} \left( a_k - \bacc_k - \eta^{ad}_k \right) \Delta t,
		\\
		& \Delta \position_{ij} \triangleq \sum_{k=i}^{j-1} \left[ \Delta \vel_{ik}\Delta t + \frac{1}{2}\Delta \R_{ik}(a_k - \bacc_k -\eta_k^{ad} \Delta t^2 ) \right],
	\end{split}
\end{align}
where $\Delta \R_{ij}$, $\Delta \vel_{ij}$ and $\Delta \position_{ij}$ are the rotation, velocity and translation differences between times $t_i$ and $t_j$ denoted by $\Delta t$, $ad$ is the accelerometer standard deviation and $gd$ is the gyroscope standard deviation.

This summarized relative state is used as a motion model constraint between the previous and current state of the body. By isolating the noise terms, we get
\begin{align}
	\begin{split}
		& \Delta \R_{ij} = \R_i^T \R_j\text{Exp}(\delta\phi_{ij}),
		\\
		& \Delta \vel_{ij} = \R_i^T(\vel_j - \vel_i - \text{g}\Delta t_{ij}) + \delta \vel_{ij},
		\\
		& \Delta \position_{ij} = \R_i^T(\position_j - \position_i - \vel_i \Delta t_{ij} - \frac{1}{2}\text{g}\Delta t_{ij}^2) + \delta \position_{ij}.
	\end{split}
\end{align}
with $g$ being the value of Earth's gravity, and $\delta \phi_{ij}$, $\delta \vel_{ij}$ and $\delta \position_{ij}$ being the rotation, velocity and translation increments between times $t_i$ and $t_j$ due to the noise.
For details on noise and covariance propagation, please refer to~\cite{Forster17tro}.

%% file: 04_approach.tex
%%%%%%%%%%%%%%%%%%%%%%%%%%%%%%%%%%%%%%%%%%%%%%%%%%%%%%%%%%%%%%%%%%%%%%%%%%%%%%%%
\section{APPROACH}

In this section, the state estimation problem is formulated and we describe the overall approach modeled by the factor graph. We assume the IMU is rigidly attached to the base link, and that the robot either has force/tactile sensors on its feet or has some contact detection mechanism~\cite{Jenelten19ral} to give us foot contact information.

\subsection{Problem Formulation}

The state variables are denoted by the pose of the base link $P^W_B$ in some arbitrary world coordinate frame $W$, and its linear velocity $V_B$. Each foot link has its pose at contact in the world frame denoted by $P^W_{L_i}$, where $i \in \{N\}$ is the index of the corresponding leg for $N$ legs. We denote the bias of the accelerometer and the gyroscope as $\bacc$ and $\bgyr$ respectively. This gives the robot state at time $t_i$ as
\begin{equation}
	\label{eq:state}
    x_i \dot{=} [P^W_{Bi}, V_B, \bgyr, \bacc] \in \textbf{SE}(3) \times \mathbb{R}^9.
\end{equation}

Between two time steps $i$ and $j$, we denote the IMU measurements as $(\omega_{i,j}, a_{i, j})$, the joint measurements as $J_{i,j}$ and the contact states for all legs as $C_{i, j}$. The set of all measurements up to time-step $k$ is denoted by
\begin{equation}
\label{eq:measurement-set}
Z^k \dot{=} \{ \omega^k, a^k, J^k, C^k \}.
\end{equation}

Given the measurements from the IMU, joint encoders, and contact sensors, we frame our problem as probabilistic inference. By assuming the measurements are conditionally independent and the noise on the sensors are zero mean additive Gaussians, the posterior probability of the state estimation problem can be stated as
\begin{align*}
	\label{eq:factor-graph}
    & \phi(X^K) \propto p(X^K \vert Z^k) \\
	& \dot{=} \phi(x_0) \prod_{k = 1}^{K-1} \phi^{\text{IMU}}(x_k, x_{k-1}) \prod_{j=1}^{L} \phi^{FK}(x_k, l_{k,j}) \phi^{CP}(l_{k,j}, c_i)
\end{align*}
where $\phi^{IMU}$ are the preintegrated IMU factors, $\phi^{FK}$ are the Forward Kinematics factors, and $\phi^{CP}$ are the factors equating the leg end-effector to the same contact point through a stance phase.

\begin{figure}[t!]
	\vspace{0.5em}
	\captionsetup{font=footnotesize}
	\centering
	\includegraphics[width=7cm,height=6cm,trim={1.2cm 6.4cm 11.7cm 13.69cm},clip]{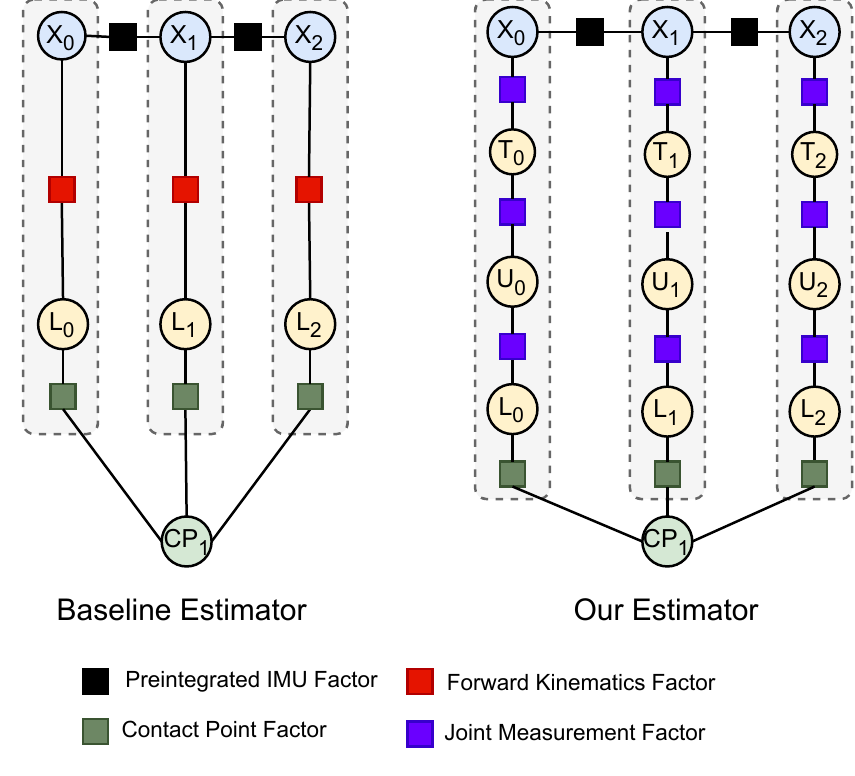}
	\caption{The factor graphs for the baseline estimator and our proposed state estimator. We propose modeling each joint in the leg's kinematic chain to apply a noise model, removing the need for noise assumptions on the forward kinematics between the base and the end link. Here, in the case of the A1, \textit{B} represents the base link, \textit{T} the thigh link, \textit{U} the upper leg link, and \textit{L} the lower leg link. The variable for the contact point is denoted by \textit{CP}.}
	\label{fig:factor_graphs}
	\vspace{-1.5em}
\end{figure}

\subsection{Preintegrated IMU Measurements}
In the same vein as~\cite{Bloesch13iros_se_slippery_uneven,Bloesch12rss_se_imu_leg_kinematics}, we use the measurement data from the IMU located at the base of the robot. We perform preintegration of the IMU measurements~\cite{Forster17tro} and add a preintegrated IMU factor at the specified estimation rate (50 Hz for quadrupeds, 100 Hz for the humanoid). The estimation rate for factor addition can be updated depending on the use case, allowing us to recover estimates at a rate suitable for use in trajectory controllers. A smoothing based approach also helps avoid drift due to accumulated linearization errors, commonly seen in filtering based approaches.

\subsection{Kinodynamic Characterization with Factor Graphs}
The robot kinematic state can be modeled as a factor graph as the links and joints follow a similar bipartite structure~\cite{Xie20arxiv_DFGP}. A robot's kinematic model can be specified as $\textit{R} = \{J, L, \mathfrak{E}\}$ which is also a bipartite graph between joints $J$ and links $L$, with the edges $\mathfrak{E}$ specifying the connections. Thus, for each joint we can specify a joint measurement factor $\phi^{FK}_j$ which takes the joint angle as a measurement and provides the relationship between the connecting links $\{ L_j, L_{j+1}i \}$.

An additional, non-obvious, benefit of the factor graph framework is the potential for extensibility. One can add additional factors to incorporate prior knowledge as well as encode additional properties of the robot such as its dynamics (\eg inertial matrix, friction cone). This allows for potentially also estimating the centroidal dynamics of the robot~\cite{Dai14humanoids} which we leave for future work.

\subsection{Forward Kinematics at Contact}
At each point a leg is in contact, we add the sub-graph to model the kinematic chain from the base of the robot to the link in contact. This allows us to probabilistically model the interactions between the links' noise models, rather than assume gaussian noise over the forward kinematics. In contrast to~\cite{Bloesch12rss_se_imu_leg_kinematics, Hartley18icra_contact_factors} which uses a single forward kinematics constraint between the base link and the foot link, we add multiple pose factors which enforce a transformation between each joint's parent and child links. Each factor takes as measurement the angle of the joint as recorded by a joint encoder or joint-level IMU. By using a minimal representation of the joint's screw axis as described in~\citet{Lynch17book_robotics}, we are able to ensure an efficient representation despite the addition of more factors. Fig~\ref{fig:factor_graphs} provides a visualization.

For $J^f$ joints in the kinematic chain for each leg $f$ in the set of legs $F$, each joint encoder measurement $z_{k,j}$, and contact $c_{k,f}$ at timestep $k$, we get the following factorization of the posterior $p(X_k, L_{k,1:F} | z_{k,1:J}, c_{k,1:F})$ on the base state $X_k$ (equal to the link pose $L^1_k$) and all leg links $L_{k,1:F}$:
\begin{equation}
	\label{eqn:contact_joint_prob}
	\begin{aligned}
		& p(X_k, L_{k,1:F} | z_{k,1:J}, c_{k,1:F}) \propto \\
		& \prod_{f \in F} \phi^{FK}(X_k, L^{2}_{k,f}|z_{k, 1}, c_{k,f}) \prod^{J^f}_{j=2} \phi^{FK}(L^j_{k,f}, L^{j+1}_{k,f} | z_{k,j}, c_{k,f})
	\end{aligned}
\end{equation}

\subsection{Contact Points as Landmarks}
In order to constrain the foot positions and estimate the IMU biases, we also estimate the contact points as being consistent across the leg's stance phase.
By viewing the state estimation problem as a SLAM problem, we consider the contact points as landmarks to be estimated and use techniques from SLAM (e.g. data association) directly.

The Forward Kinematics ($\mathrm{FK}$) provides a mapping between the base link of the robot and the foot contact point, given that the foot is in contact. We specify a foot is in contact with point $m \in \{M\}$ at time-step $k$ using the factor $\phi^{CP}(X_k, C_{im})$ which minimizes the distance between the contact point $m$ in the environment and the foot link $i \in F$
\begin{equation}
\label{eq:contact-constraint}
    \begin{aligned}
        & \phi^{CP}(X_k, C_{im}) = \mathrm{g}^{CP}(X_k, C_{im}) = \Vert \mathrm{FK}(X_k, L_i) - C_{im} \Vert
    \end{aligned}
\end{equation}
Depending on the type of foot, either Point or Flat, we can constrain just the foot translation or the foot pose~\cite{Rotella14iros}.

\subsection{Maximum A Posteriori Estimation}
Given a linearization point $X$, we linearize our factor graph and optimize for the negative log-likelihood to give us the best estimates of the state trajectory using the Levenberg-Marquardt optimization method
\begin{equation}
	\begin{aligned}
		& X^{*} = \arg \min_{X} \bigg( \Vert o(X_0) \Vert^2_{\Sigma_0} + \sum_{k=1}^{K-1}\Vert \mathrm{f}^\mathrm{IMU}(X_k, X_{k-1}) \Vert^2_{\Sigma_\mathrm{IMU}}
		\\
		& + \sum_{k=0}^{K-1} \sum_{f}^{F}\Vert \mathrm{h}^\mathrm{FK}(X_k, L_{k,f}^2) \Vert^2_{\Sigma_{\mathrm{FK}}}\sum_{i=2}^N \Vert \mathrm{h}^\mathrm{FK}(L^{i}_{k,f}, L^{i+1}_{k,f}) \Vert^2_{\Sigma_{\mathrm{FK}}}
		\\
		& + \sum_{k=0}^{K-1}\sum_{i}^{F} \sum_{m}^{M} \Vert \mathrm{g}^\mathrm{CP}(X_k, C_{im}) \Vert^2_{\Sigma_\mathrm{CP}} \bigg)
	\end{aligned}
\end{equation}
where $o(X_0)$ is a prior on the first state, $\mathrm{f}^{\mathrm{IMU}}$, $\mathrm{h}^\mathrm{FK}$ and $\mathrm{g}^\mathrm{CP}$ are the linearized binary preintegrated IMU error, forward kinematics error for a parent-child link pair, and the contact error for foot $i$, respectively. $\Sigma_0, \Sigma_\mathrm{IMU}, \Sigma_\mathrm{FK}, \Sigma_\mathrm{CP}$ are the respective noise model covariances.

We compute a good linearization point using the preintegrated IMU measurements and the joint angle measurements, giving us initial values for the state and the contact points.

%
%\subsection{Hybrid Dynamics Model}
%
%For our approach, we propose a hybrid dynamics model which includes 3 different states of foot contact. These 3 states are:
%
%\begin{enumerate}
%	\item Fixed Contact (C): The foot is rigidly planted on the surface and the contact is not moving.
%	\item Slipping Contact (S): The foot is touch the surface but is experiencing slip.
%	\item Flight (F): The foot is not touching a contact surface.
%\end{enumerate}
%
%\subsection{Friction Cones for Hybrid State Mapping}
%
%To convert the discrete aspects of the hybrid dynamics model into a continuous optimization problem, we use the formulation for fuzzy contact weights in~\cite{Gassmann05icra_walking_localization}.
%2. Humanoid dynamics via the URDF
%4. 

%% file: 05_results.tex
\section{Experimental Results}

%\subsection{Datasets}

%\subsection{Metrics}

To evaluate our approach, we use common metrics used for SLAM\cite{Sturm12iros, Kuemmerle09ar} such as Absolute Pose Error and Relative Pose Error for investigating the global and local consistency of the trajectories respectively. We make use of the \textit{evo} software package~\cite{Grupp17evo} to provide us with convenient functions for the metrics. The Absolute Pose Error (\textbf{APE}) is defined as the absolute relative pose between two poses, the reference pose $P_i$ and the estimated pose $\hat{P}_i$ at timestamp $i$
\begin{equation}
	\label{eq:ape}
	APE_i = \Vert (P_i^{-1} \hat{P}_i) - I_{4\times4} \Vert_F
\end{equation}
The Relative Pose Error (\textbf{RPE}) compares the relative poses along the estimated and reference trajectories and is computed over $1$ second
\begin{equation}
	\label{eq:rpe}
	RPE_{i,j} = \Vert [(P_i^{-1}P_j)^{-1}(\hat{P}_i^{-1}\hat{P}_j)] - I_{4\times4} \Vert_F
\end{equation}

To compute the metrics, we first have to align the trajectories, which we do with a novel scheme for legged robots. Trajectory alignment ensures that the error being computed is not dominated by any particular unobservable gauge freedom, \eg scale in traditional SLAM systems. In our case, we have 4 degrees of gauge freedom, namely the translation (contributing 3 degrees of freedom) and the yaw $\gamma$, all of which are directly unobservable for legged robots.

The trajectory alignment is done in two steps, the $(x, y, \gamma)$ and the $z$ axis. Aligning the $(x, y, \gamma)$ is a straightforward $SE(2)$ transformation calculated by computing the best alignment between the $(x, y)$ points of the trajectories. For aligning the $z$ axis, we subtract the mean of the $z$ values of the trajectory translations from each trajectory independently.

\renewcommand{\arraystretch}{1.1}

\begin{table}[h]
	\vspace{1em}
	\centering
	\captionsetup{font=footnotesize}
	\caption{Experimental Results on Quadrupeds in both simulation (A1) and hardware (Anymal C). We report the Root Mean Square Error (RMSE) for Absolute Pose Error (APE) and Relative Pose Error (RPE) over 1 second.}
	\begin{tabular}{|c|cc|cc|}
		\hline
		& \multicolumn{2}{c}{\textbf{Baseline Estimator}} & \multicolumn{2}{|c|}{\textbf{Our Approach}} \\ \hline \hline
		\multicolumn{1}{|c|}{Trajectory} & APE & RPE & APE & RPE     \\ \hline \hline
		\multicolumn{1}{|c|}{A1 Straight} &  0.207 & 0.0063  & 0.205 & 0.0063 \\ \hline
		\multicolumn{1}{|c|}{A1 Diagonal} &  0.136 & 0.0064  &  0.135 & 0.0064 \\ \hline
		\multicolumn{1}{|c|}{A1 Turn}     &  0.190 & 0.0053  &  0.181 & 0.0049 \\ \hline
		\multicolumn{1}{|c|}{A1 Zig-Zag}  &  0.109 & 0.0062  &  0.108 & 0.0063 \\ \hline
		\multicolumn{1}{|c|}{Anymal C}    &  0.307 & 0.0115  & 0.173 & 0.0046 \\ \hline
	\end{tabular}
	\label{table:quadruped-results}
\end{table}

\begin{table}[h]
	\centering
	\captionsetup{font=footnotesize}
	\caption{Experimental Results on a DRC-era Atlas humanoid over 5 different test runs. Similar to Table~\ref{table:quadruped-results}, we report the Root Mean Square Error (RMSE) for Absolute Pose Error (APE) and Relative Pose Error (RPE) over 1 second for each trajectory.}
	\begin{tabular}{|c|cc|cc|}
		\hline
		& \multicolumn{2}{c|}{\textbf{Baseline Estimator}} & \multicolumn{2}{c|}{\textbf{Our Approach}} \\ \hline \hline
		\multicolumn{1}{|c|}{Trajectory} & APE & RPE & APE & RPE \\ \hline \hline
		\multicolumn{1}{|c|}{Atlas Run 1} & 0.076 & 0.0201 & 0.067 & 0.0182 \\ \hline
		\multicolumn{1}{|c|}{Atlas Run 2} & 0.192 & 0.0339 & 0.141 & 0.0277 \\ \hline
		\multicolumn{1}{|c|}{Atlas Run 3} & 0.198 & 0.0313 & 0.098 & 0.0201 \\ \hline
		\multicolumn{1}{|c|}{Atlas Run 4} & 0.154 & 0.0267 & 0.114 & 0.0189 \\ \hline
		\multicolumn{1}{|c|}{Atlas Run 5} & 0.292 & 0.0324 & 0.271 & 0.0260 \\ \hline
	\end{tabular}
	\label{table:humanoid-results}
	\vspace{-1.5em}
\end{table}

We evaluate our estimator in simulation to show correctness and establish the strong baseline. We use the pybullet based simulation environment provided by~\cite{Peng20rss}. We generate four different trajectories in simulation which we refer to as straight, diagonal, turn, and zig-zag. In the simulation, along with the ground truth state values, we collect the joint angles for all the joints in the robot and the contact state of each leg. We add realistic noise to the simulated measurements, \ie from the IMU  and the joint angles. We execute both the baseline estimator and our estimator on the different trajectories.

The baseline estimator used is a factor-graph based implementation of the proprioceptive state estimator described by~\cite{Bloesch12rss_se_imu_leg_kinematics}. We choose this baseline since it is an estimator that is widely deployed in many real-world robots~\cite{Hutter12clawar_starleth,Bledt18iros,Kuindersma16ar_atlas}. By implementing a factor graph based version of the original filter-based estimator, we bring forward the benefits of the original filter while also allowing for a fair comparison between the baseline and our proposed framework.

\begin{figure}[h!]
	\captionsetup{font=footnotesize}
	\centering
	\begin{subfigure}{0.35\textwidth}
		\includegraphics[width=\textwidth]{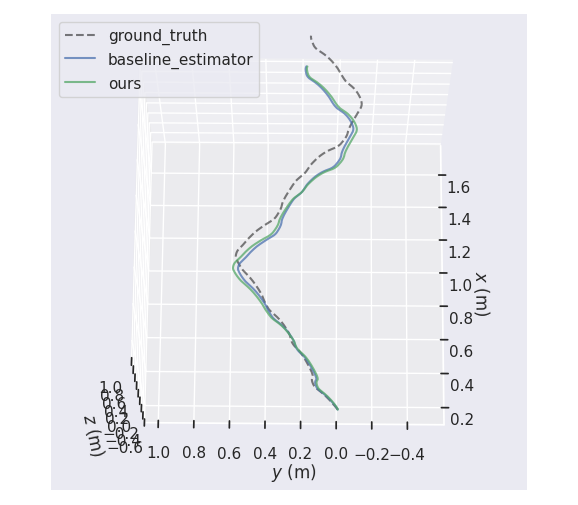}
	\end{subfigure}
	\begin{subfigure}{0.35\textwidth}
		\includegraphics[width=\textwidth,height=3.6cm]{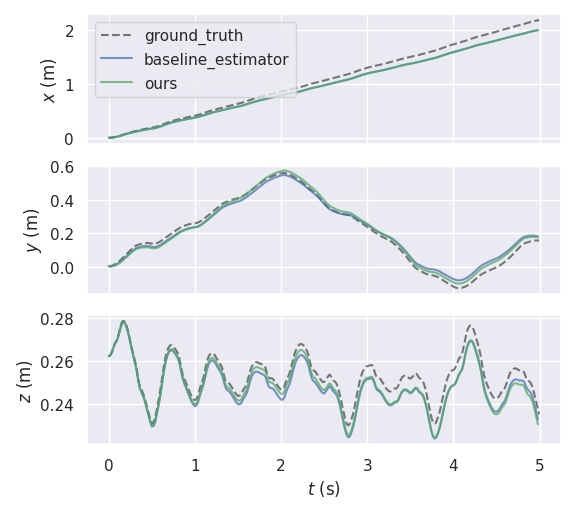}
	\end{subfigure}
	\begin{subfigure}{0.35\textwidth}
		\includegraphics[width=\textwidth,height=3.6cm]{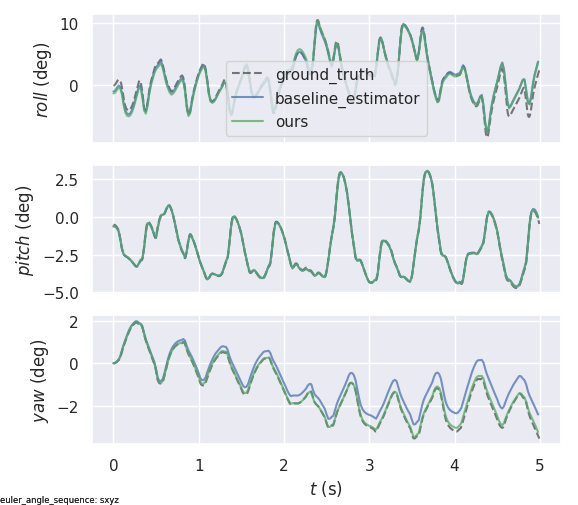}
	\end{subfigure}
	\caption{The A1 walking a zig-zag trajectory in simulation. Qualitative comparisons of (a) the trajectories, (b) the translations, and (c) the rotations, verifying the performance of our state estimator in ideal conditions.}
	\label{fig:zigzag}
\end{figure}

%\begin{figure*}[h!]
%	\captionsetup{font=footnotesize}
%	\centering
%	\begin{subfigure}{0.4\textwidth}
%		\includegraphics[width=\textwidth]{images/turn/trajectory.png}
%		\vspace{0.01em}
%	\end{subfigure}
%	\begin{subfigure}{0.23\textwidth}
%		\includegraphics[width=\textwidth]{images/turn/xyz.png}
%	\end{subfigure}
%	\begin{subfigure}{0.23\textwidth}
%		\includegraphics[width=\textwidth]{images/turn/rpy.png}
%	\end{subfigure}
%
%	\caption{Trajectory with sharp turn. We show qualitative comparisons of (a) the trajectories, (b) the translations, and (c) the rotations.}
%	\label{fig:turn}
%\end{figure*}

%\subsection{Hardware}

Finally, we demonstrate the efficacy of our estimator on robot hardware. We run our estimator on a sample trajectory of the ANYbotics Anymal C robot and a Boston Dynamics Atlas robot. We track the trajectory of the Anymal C base using a Vicon optical tracking system, akin to~\cite{Wisth21tro_vilens}.
% and the pelvis of the Atlas with an OptiTrack system
For the Atlas, we use the estimates provided by the well-tuned built-in state estimator as a pseudo-ground-truth trajectory.

\begin{figure}[h!]
	\captionsetup{font=footnotesize}
	\centering
	\begin{subfigure}{0.35\textwidth}
		\includegraphics[width=\textwidth]{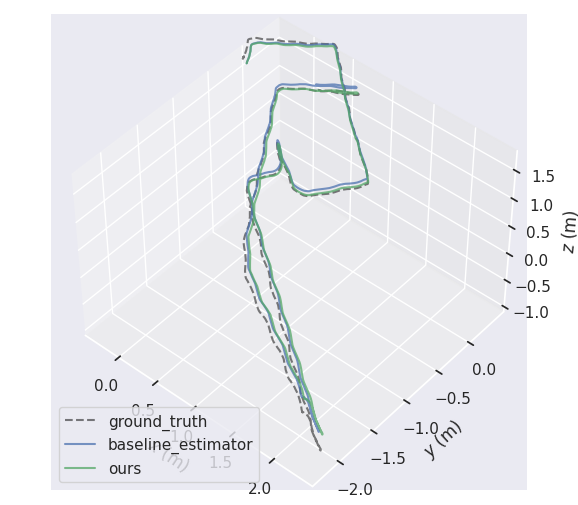}
		\vspace{0.01em}
	\end{subfigure}
	\begin{subfigure}{0.35\textwidth}
		\includegraphics[width=\textwidth]{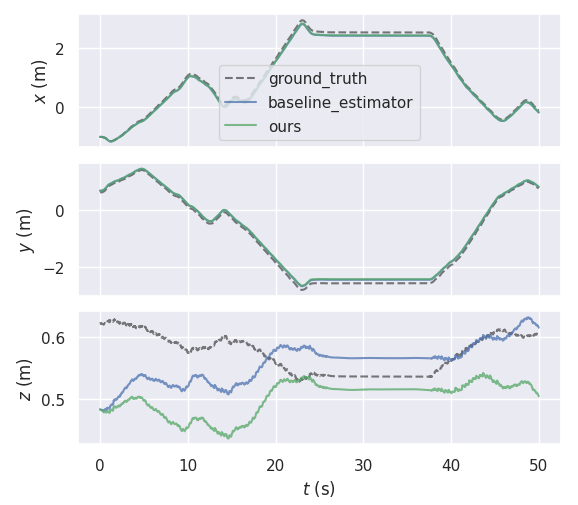}
	\end{subfigure}
	\begin{subfigure}{0.35\textwidth}
		\includegraphics[width=\textwidth]{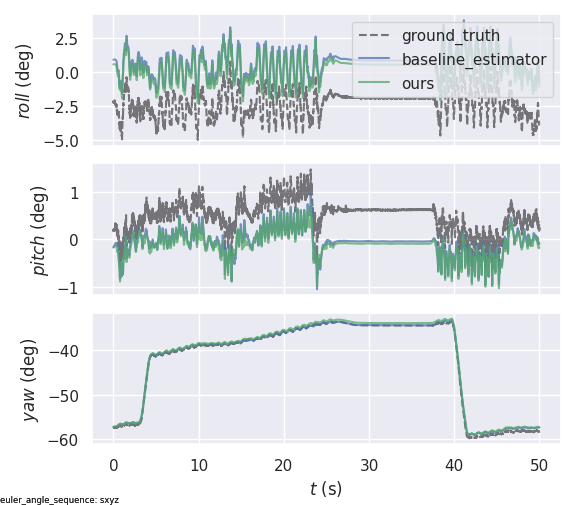}
	\end{subfigure}
	
	\caption{State estimation for the Anymal C robot walking a non-trivial trajectory. The ground truth trajectory was gathered using motion capture technology in a lab environment.}
	\label{fig:anymalc_traj}
\end{figure}

\begin{figure}[h!]
	\captionsetup{font=footnotesize}
	\centering
	\begin{subfigure}{0.35\textwidth}
		\includegraphics[width=\textwidth]{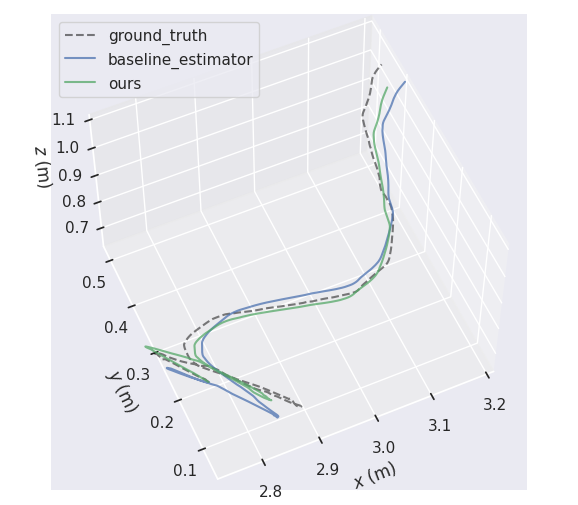}
		\vspace{0.01em}
	\end{subfigure}
	\begin{subfigure}{0.35\textwidth}
		\includegraphics[width=\textwidth]{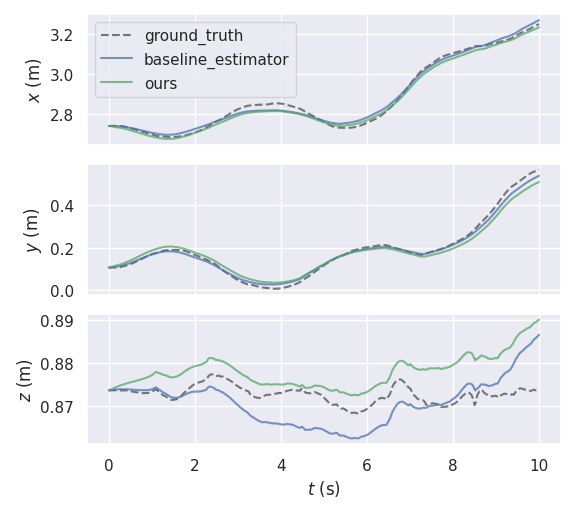}
	\end{subfigure}
	\begin{subfigure}{0.35\textwidth}
		\includegraphics[width=\textwidth]{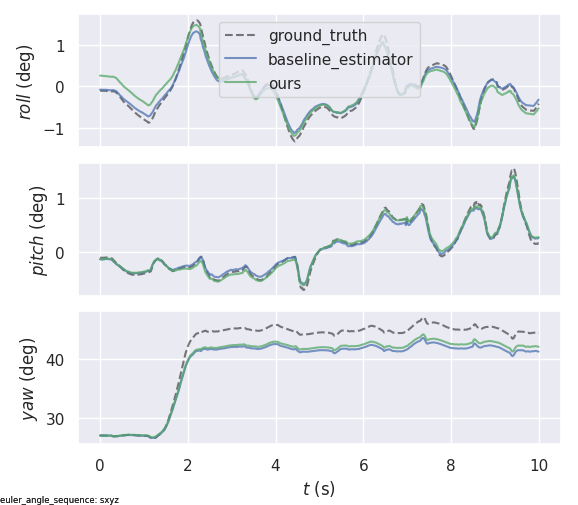}
	\end{subfigure}
	
	\caption{State estimation results for the Atlas robot. Both rotation and translation estimates are equivalent or outperform the baseline estimator.}
	\label{fig:atlas_traj}
\end{figure}

%We report the results in Table \ref{table:quadruped-results} and show a qualitative example result in figure~\ref{fig:zigzag}.
%~\ref{fig:straight},~\ref{fig:diagonal},and~\ref{fig:turn}.
Quantitative results for the A1 simulation and the Anymal C in Table~\ref{table:quadruped-results} and for the Atlas in Table~\ref{table:humanoid-results}, show our state estimator's superior performance over a variety of trajectories for both platforms. As seen from the reported results, our state estimator outperforms the baseline estimator in APE and RPE metrics by 27.62\% and 28.75\% on average respectively. The RPE improvement illustrates improved drift handling over 1 second intervals, and in particular, for 5 different trajectories of the Atlas humanoid, we show improved state estimates consistently, demonstrating both the accuracy and the capability of our approach.
Sample trajectories are displayed in Figures~\ref{fig:zigzag},~\ref{fig:anymalc_traj},~\ref{fig:atlas_traj}.

%% file: 06_conclusion.tex
\section{CONCLUSION}
In this work, we presented a factor-graph based framework for proprioceptive legged robot state estimation. By viewing the robot's kinematic structure as a bipartite graph, we are able to model it using a factor graph together with the IMU preintegration factors. This in-turn allows for constraining the IMU biases for accurate and efficient proprioceptive state estimation. We demonstrate the efficacy and generalizability of our approach on both humanoids and quadrupeds, showing improved performance when compared to a strong baseline in both simulation and robot hardware.

Our framework is extensible and allows for asynchronous data handling. In future work, we hope to examine the effects of new factors based on force-torque measurements and constraints based on slip-estimation. Our hope is that this work serves as a stepping-stone towards more widespread deployment of legged robots for various applications.

\section*{ACKNOWLEDGMENTS}
We thank Russell Buchanan and Bhavyansh Mishra for help with the Anymal C robot data and the Atlas robot data, respectively. We would also like to thank Jerry Pratt, Maurice Fallon and Fan Jiang for helpful discussions and advice.